\documentclass[lettersize,journal]{IEEEtran}

\usepackage[english]{babel}

\usepackage{graphicx}

\usepackage[utf8]{inputenc}

\usepackage{mathbbol}

\usepackage{esvect}

\usepackage[percent]{overpic}

\usepackage{tabu}

\usepackage{multirow}

\usepackage[overload]{empheq}

\usepackage{bm}

\usepackage{xcolor}
\definecolor{dg}{rgb}{0.1, 0.6, 0.2}       
\definecolor{b}{rgb}{0.0, 0.0, 1}          
\definecolor{lightblue}{rgb}{0.0, 0.7, 0.95}    
\definecolor{lightgreen}{rgb}{0.0, 0.7, 0.3}    

\usepackage{epsfig}
\usepackage{float}
\usepackage{color}
\usepackage{url}

\usepackage{algorithm,algpseudocode}
\newcounter{phase}[algorithm]
\newlength{\phaserulewidth}
\newcommand{\setphaserulewidth}{\setlength{\phaserulewidth}}

\setphaserulewidth{.7pt}

\usepackage{amssymb, amsfonts}
\usepackage{amsmath}
\usepackage{mathrsfs}

\usepackage{amsthm}
\usepackage{mathtools}

\usepackage{enumitem}       
\setenumerate[enumerate]{align=left}

\usepackage{graphicx}
\usepackage{subcaption}

\usepackage{cite}

\makeatletter
\let\NAT@parse\undefined
\makeatother
\usepackage{hyperref} 
\usepackage{footmisc}   

\usepackage{balance}

\hypersetup{
    colorlinks=true,
    linkcolor=blue,
    filecolor=blue,      
    urlcolor=blue,
    citecolor=blue,
}

\title{Cooperative Aerial Robot Inspection Challenge: A Benchmark for Heterogeneous Multi-UAV Planning and Lessons Learned}
\author{
        Muqing Cao$^{1*}$,
        Thien-Minh Nguyen$^{2*\dag}$,
        Shenghai Yuan$^2$,
        Andreas Anastasiou$^3$, Angelos Zacharia$^3$, Savvas Papaioannou$^3$, Panayiotis Kolios$^3$, Christos G. Panayiotou$^3$, Marios M. Polycarpou$^3$, 
        Xinhang Xu$^2$,
        \\
        Mingjie Zhang$^4$,
        Fei Gao$^5$,
        Boyu Zhou$^4$, 
        Ben M. Chen$^6$,
        Lihua Xie$^2$
\thanks{$^1$ Robotics Institute, Carnegie Mellon University. $^2$ School of Electrical and Electronic Engineering, Nanyang Technological University. $^3$ KIOS Research and Innovation Center of Excellence, Department of Electrical and Computer Engineering, University of Cyprus. $^4$ School of Artificial Intelligence, Sun Yat-sen University. $^5$ Zhejiang University. $^6$ Department of Mechanical and Automation Engineering, Chinese University of Hong Kong.}
\thanks{$^*$ Equal contribution. $^\dag$ Corresponding author: Thien-Minh Nguyen (thienminh.npn@ieee.org)}
}
\date{}

\begin{document}
\maketitle

\begin{abstract}
We propose the Cooperative Aerial Robot Inspection Challenge (CARIC), a simulation-based benchmark for motion planning algorithms in heterogeneous multi-UAV systems. CARIC features UAV teams with complementary sensors, realistic constraints, and evaluation metrics prioritizing inspection quality and efficiency. It offers a ready-to-use perception-control software stack and diverse scenarios to support the development and evaluation of task allocation and motion planning algorithms.
Competitions using CARIC were held at IEEE CDC 2023 and the IROS 2024 Workshop on Multi-Robot Perception and Navigation, attracting innovative solutions from research teams worldwide. This paper examines the top three teams from CDC 2023, analyzing their exploration, inspection, and task allocation strategies while drawing insights into their performance across scenarios.
The results highlight the task's complexity and suggest promising directions for future research in cooperative multi-UAV systems.

\end{abstract}


\section{Introduction}
\begin{figure}
    \centering
    \includegraphics[width=1\linewidth]{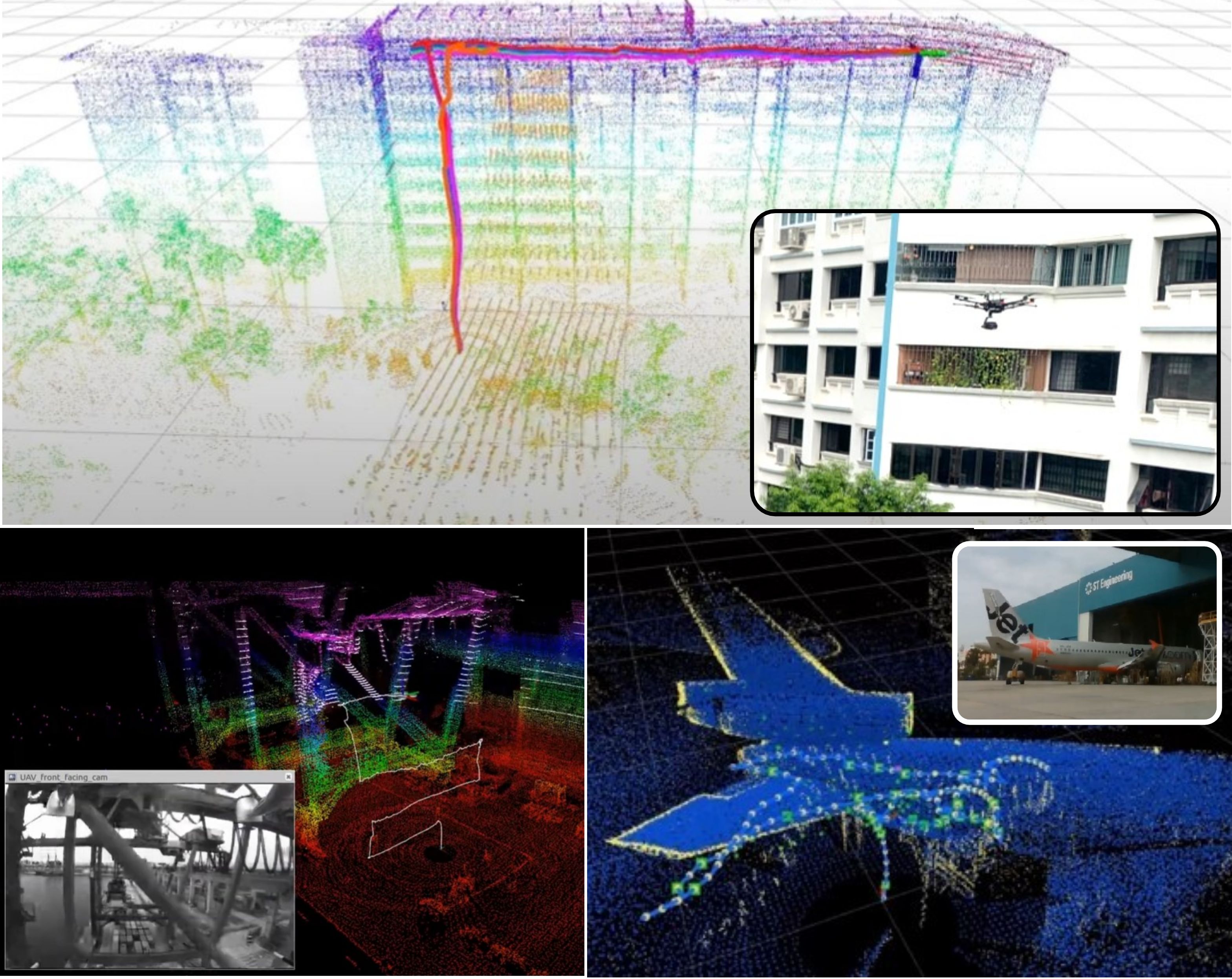}
    \caption{Examples of aerial inspection mission for building, crane, aircraft in our previous projects.}
    \label{fig: inspection example}
\end{figure}

Aerial robots have become widely adopted for the inspection of complex structures, such as buildings, cranes, bridges, and airplanes \cite{bircher2015structural, ZHANG2022bridge, Saha2023aircraft}; see Figure \ref{fig: inspection example}. Existing commercial systems mainly rely on human operators who manually pilot UAVs. While effective, this approach is labor-intensive, time-consuming, and prone to human error. 
As a result, the robotics community has increasingly focused on autonomous structural inspection using UAVs to improve efficiency and reliability.

However, most autonomous inspection solutions rely on path planning algorithms that require a detailed prior model of the structure to be inspected \cite{bircher2015structural, MANSOURI2018cooperativeCEP, Huang2023BIM}. Obtaining such models often requires a separate data collection phase, which adds to operational complexity and cost. Efforts to perform online modeling and inspection using onboard cameras have been explored \cite{bircher2018receding, Papachristos2019, song2022view}, but cameras are generally inefficient for mapping due to their limited field of view and sensitivity to lighting conditions. LiDAR sensors, on the other hand, are highly effective in capturing dense 3D information over large areas quickly, making them ideal for creating a structural prior for inspection.

Given the cost and payload limitations of equipping all UAVs with LiDAR, heterogeneous teams of drones, some equipped with both LiDAR and camera for mapping and inspection and others with only cameras, offer a promising solution. This setup allows for fast and efficient online modeling and inspection, leveraging the strengths of both sensors to balance cost-effectiveness and operational efficiency.
However, designing effective strategies for task allocation, path planning, and coordination in heterogeneous teams is a complex problem, especially with practical limitations such as battery and communication constraints. 
While single-robot or homogeneous multi-robot inspection systems have been studied extensively \cite{ivic2023multi, jing2020multi, Hardouin2023reconstruction}, heterogeneous multi-UAV inspection remains underexplored, leaving critical gaps in research and development.

To address these gaps, we introduce the Cooperative Aerial Robot Inspection Challenge (CARIC) to stimulate innovative solutions for heterogeneous UAV inspection and provide a platform for researchers and practitioners to test their multi-robot inspection planning algorithms. The challenge offers a lightweight and ready-to-use simulation framework, featuring a team of high-cost mapping and exploration drones and low-cost inspection drones tasked with inspecting simulated industrial sites under conditions mimicking real-world operations.

The primary challenge lies in designing efficient multi-robot task allocation and motion planning algorithms that can capture high-quality structural images while adhering to practical constraints. These constraints include: \begin{itemize} \item Lack of prior model: Instead of providing a detailed map of the inspection environment, only bounding boxes of the structures to be inspected are available. This reflects realistic scenarios where generating a detailed map is resource-intensive, but bounding boxes can be estimated from a 2D map and structure height. \item Realistic evaluation metrics: The inspection performance is assessed based on image quality metrics, including blurriness and resolution, to ensure the captured data meets standards for structural analysis. \item Decentralized computation and communication constraints: UAVs communicate only when within line of sight, requiring algorithms to account for intermittent connectivity and localized decision-making. \end{itemize}

The simulation framework, including source code and detailed instructions, is publicly available at \url{https://ntu-aris.github.io/caric}. The first CARIC competition was held in conjunction with the IEEE Conference on Decision and Control (CDC) 2023.  A video summary of this event can be viewed at \url{https://youtu.be/8u5hj2oZ-VY}.
The second CARIC competition was held with the IROS 2024 Workshop on Multi-Robot Perception and Navigation Challenges in Logistics and Inspection Tasks.
The competitions showcased the diverse approaches of the participating teams to the multi-UAV inspection problem.
In this paper, we present the approaches of the top three teams of the first CARIC competition and discuss key insights and lessons learned to inspire further advancements in multi-UAV inspection research.
The key lessons reveal that achieving efficient multi-UAV inspection requires careful integration of exploration and inspection phases, accurate workload estimation for task allocation, and robust handling of communication and path-planning failures to maintain operational consistency.

\section{Benchmark Overview}
A fleet of UAVs is tasked with inspecting infrastructures such as building facades, cranes, and airplanes by capturing images of the surfaces and looking for locations with a high risk of defects.
Unlike conventional approaches that rely on detailed prior models, the UAVs are provided only with bounding boxes that encapsulate the structures. 
Therefore, the UAVs should explore the unknown volume to obtain the details of the surfaces to be inspected. 
During the inspection process, some interest points on the surface are identified as locations with high vulnerability and, therefore, need to be closely inspected with high-resolution images.
The objective of the multi-UAV system is to achieve the highest possible inspection score by capturing as many interest points as possible with the highest possible score for each point.

\begin{figure}
    \centering
    \begin{overpic}[width=\linewidth,
                   ]{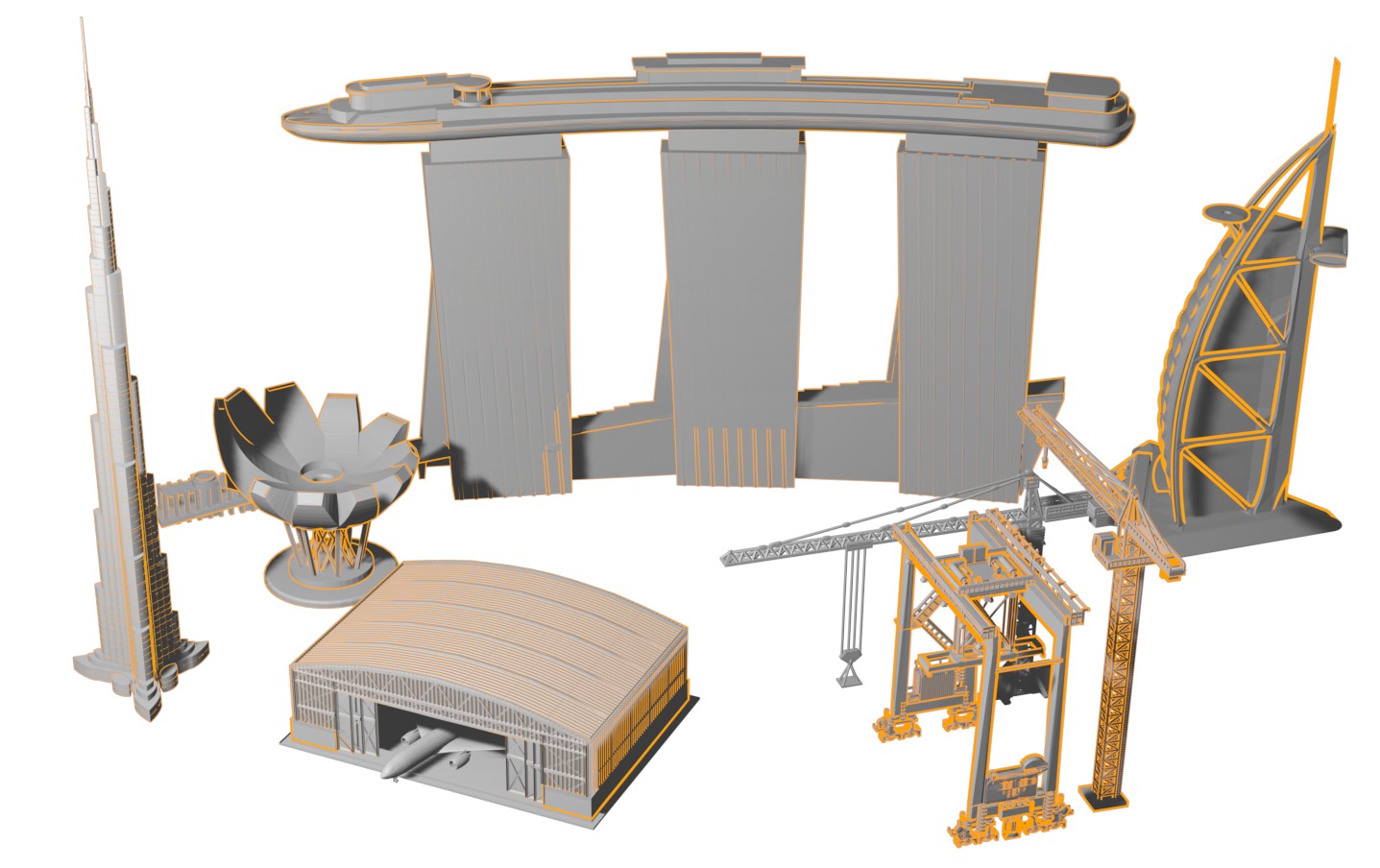}
    \put(30.00, 59.00){mbs}
    \put(20.00, 03.00){hangar}
    \put(06.00, 06.00){burj}
    \put(57.00, 03.00){crane}    
    \put(87.00, 18.00){sail}
    \end{overpic}
    \caption{Sample scenarios in CARIC.}
    \label{fig: sample scenarios}
\end{figure}
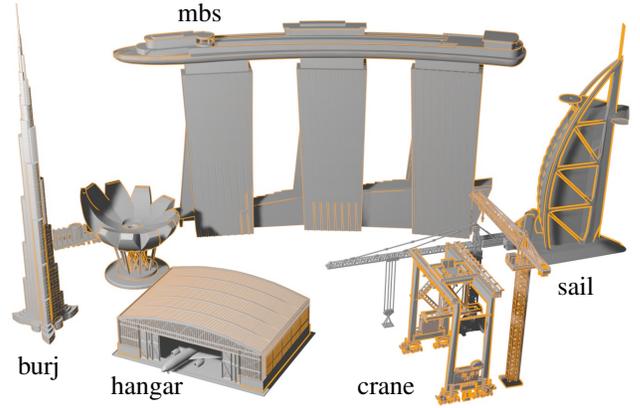

The simulation is implemented using the Gazebo simulator and RotorS \cite{furrer2016rotors}, and we provide sample scenarios for testing, as shown in Figure \ref{fig: sample scenarios}. 
The following sections define the heterogeneous UAV fleet, sensor data, control and communication requirements, and score calculation.

\subsection{UAV Fleet}

\begin{figure}
    \centering
    \begin{overpic}[width=0.35\linewidth,
                   ]{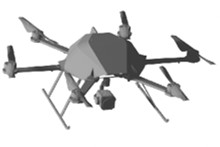}
    \put(20.00, 62.00){Photographer}
    \end{overpic}
    \begin{overpic}[width=0.5\linewidth,
                   ]{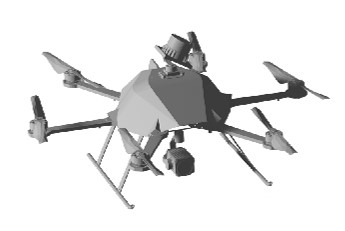}
    \put(25.00, 60.00){Explorer}
    \end{overpic}
    \caption{The heterogeneous UAV team.}
        \label{fig: fleet}
\end{figure}

The benchmark features a heterogeneous UAV team comprising of $N$ drones, categorized as explorers and photographers (Figure \ref{fig: fleet}):
\begin{itemize}
    \item Photographer: A small UAV carrying an inspection camera placed on a motorized gimbal. The main task of a photographer is to capture images of the points of interest. This setting is typical of a commercially available camera-equipped UAV, such as DJI Mavic 3.
    \item Explorer: with reference to an autonomous LiDAR-based inspection UAV \cite{Emesent}, we designed the explorer to be larger than a photographer and equipped with an inspection camera and a Lidar placed on a motorized gimbal. Besides image capturing, an explorer is capable of generating a point cloud map of the surroundings.
\end{itemize}
The team composition is flexible, with $N_p$ photographers ($N_p \in {0, \dots, N-N_e}$) and $N_e$ explorers.
Note that besides the UAV, a \textit{control station} is also present to represent human supervision and tallies the final score for the mission.

\subsection{Sensor Data}

Each UAV has access to the following basic information during simulation:
\begin{itemize}
    \item Odometry: Ground truth pose of the UAV in the global frame of reference.
    \item Gimbal pose: Relative orientation of the inspection camera with respect to the UAV body frame.
    \item 
    Inspection score: Each captured image includes a computed score that reflecting the quality of inspection at the interest points, detailed in Section \ref{sec: evaluation criteria}.
    \item LiDAR (Explorer only): A rotating LiDAR scanner generates 3D point cloud data of the surroundings at 10Hz.
\end{itemize}

Besides, we assume an underlying multi-robot localization and mapping system is available. Hence the UAVs can access certain information from \textit{the neighbours that are in its LoS}, namely \textit{odometry} and \textit{key frame point cloud}. These data ensure that the \textit{photographers} can be aware of the neighbors as well as the surroundings.

\subsection{Control and Communication}\label{subsec: control}
We implement a so-called \texttt{unicon} package in the CARIC stack to allow user to control the UAVs in two ways:
\begin{itemize}
    \item Partial control: users may send target position or velocity or acceleration and target yaw angle to the controller.
    \item Full control: users may send a command including the target position, velocity, acceleration, and yaw.
\end{itemize}
The controller node will compute lower-level control inputs to drive the UAV to the target states.
The controller will ignore a command exceeding the kinematic constraints.
For camera control, users can adjust the camera’s direction by controlling gimbal pitch and yaw, and trigger image capture via a command.

Communication between UAVs is restricted to LoS. A UAV can only send or receive messages from another UAV if they have direct visibility. We implemented this communication scheme via a so-called \texttt{ppcomrouter} node, which observes the LoS condition between objects, and relays or drops the messages accordingly.

\subsection{Evaluation Criteria} \label{sec: evaluation criteria}

The performance of the multi-UAV control strategy will be judged based on the \textbf{total score of the captured interest points received by the control station}, which is denoted as $Q$ and defined as follows:
\begin{equation} \label{eq: total score}
    Q = \sum_{i=1}^I \max_{j \in \{1,\dots N\}} q_{i,j},
\end{equation}
where $I$ is the set of interest points in the scene, $N$ is the total number of UAVs, $q_{i,j}$ is the visual inspection score of interest point $i$ by the UAV $j$.

\begin{figure}
    \centering
    \includegraphics[width=0.9\linewidth]{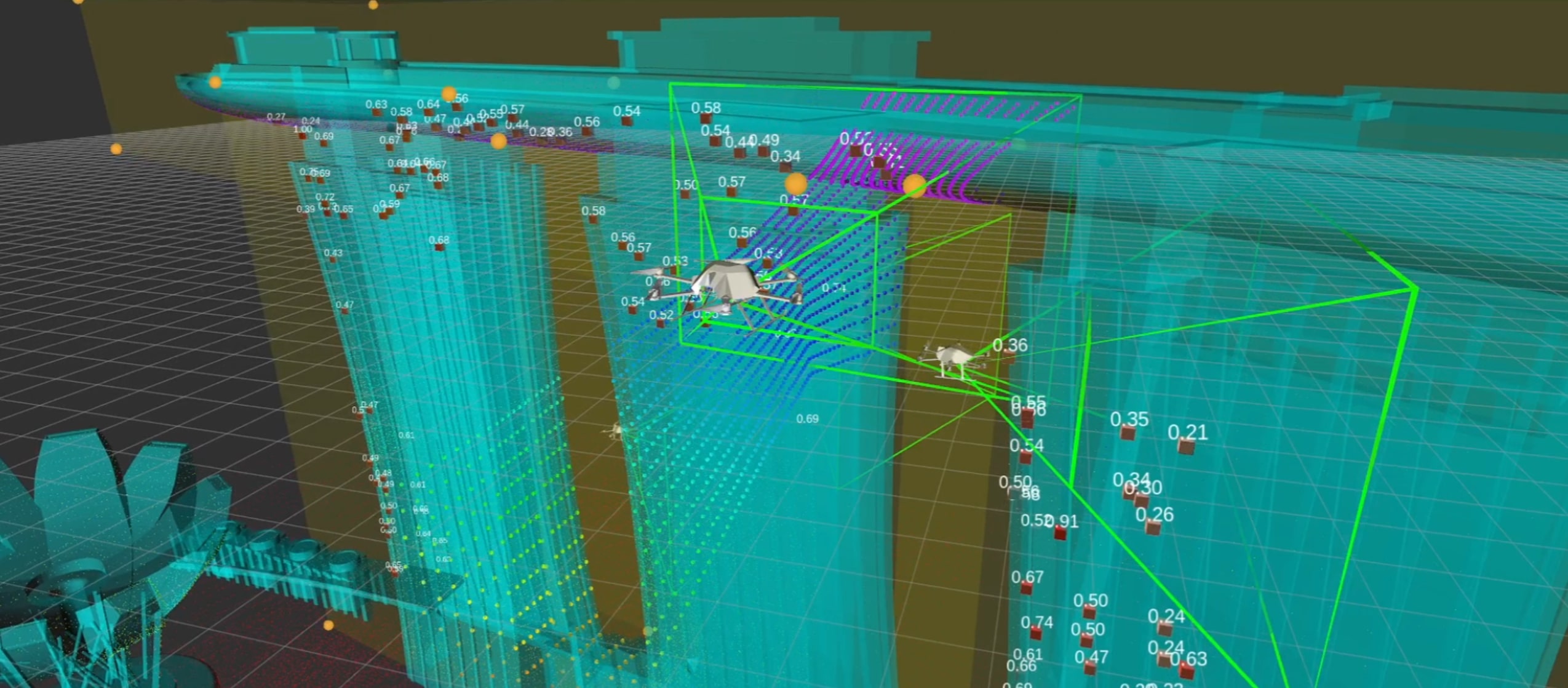}
    \caption{Example of the scores for captured interest points.}
    \label{fig: q value}
\end{figure}

The visual inspection score $q_{i,j}$ is computed as follows:
\begin{equation} \label{eq: total score2}
    q_{i,j} = \max_{k \in [0,K]} q_{i,j,k},\quad q_{i,j,k} = q_{\text{seen}}\cdot q_{\text{blur}}\cdot q_{\text{res}},
\end{equation}
where $K$ is the total mission time, $q_{i,j,k}$ is the visual inspection score of interest point $i$ obtained by UAV $j$ at a time $k$, $q_{\text{seen}}\in\{0,1\}$, $q_{\text{blur}}\in[0,1]$, $q_{\text{res}}\in[0,1]$ are the line of sight (LOS), motion blur, and resolution metrics, which are elaborated in Sections \ref{subsec: los}, \ref{subsec: blur} and \ref{subsubsec: res}, respectively. Fig. \ref{fig: q value} shows an example of the interest points (red squares) with the scores.

The value $Q$ will be calculated at the end of each mission, where only the scores of UAVs that successfully completed the mission without collision be considered. 

\subsubsection{Line of sight and field of view}\label{subsec: los}
The term $q_\text{seen}$ is a binary-valued metric value that is $1$ when the interest point falls in the field of view (FOV) of the camera, and the camera has a direct line of sight (LoS) to the interest point (not obstructed by any other objects), and 0 otherwise. 

\subsubsection{Motion blur}\label{subsec: blur}
\begin{figure}
    \centering
    \includegraphics[width=0.8\linewidth]{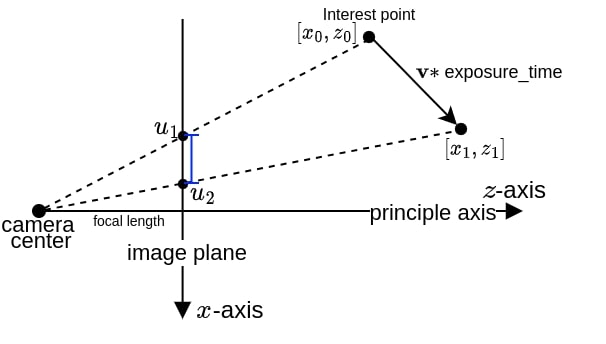}
    \caption{Illustration of Motion Blur.}
    \label{fig: blur}
\end{figure}

The motion blur metric $q_\text{blur}$ is based on the motion of the interest point during the camera exposure duration $\tau$ (a provided value) \cite{Wang2024rapid}. It can be interpreted as the number of pixels that an interest point moves across during the exposure time, i.e.:
\begin{align}
    q_\text{blur} = \min\left(\dfrac{c}{\max\left(|u_1-u_0|, |v_1-v_0|\right)}, 1.0\right), \label{eq: blur metric}
\end{align}
where $c$ is the pixel width and $\|u_1-u_0\|$, $\|v_1-v_0\|$ are the horizontal and vertical movements on the image plane that are computed by:
\begin{gather}
    u_0 = f\cdot\dfrac{x_0}{z_0},
u_1 = f\cdot\dfrac{x_1}{z_1},
v_0 = f\cdot\dfrac{y_0}{z_0}, 
v_1 = f\cdot\dfrac{z_1}{z_1},\\
[x_1,y_1,z_1]^\top = [x_0,y_0,z_0]^\top + \mathbf{v}\cdot\tau,
\end{gather}
with $f$ being the focal length, $[x_0,y_0,z_0]^\top$ the position of the interest point at the time of capture, and $[x_1,y_1,z_1]^\top$ the updated position considering the velocity of the interest point in the camera frame at the time of capture, denoted as $\mathbf{v}$. 
The calculation of this velocity requires advance kinematic analysis, and is detailed on our website \footnote{\url{https://ntu-aris.github.io/caric/docs/CARIC_motion_blur.pdf}}. 
Figure \ref{fig: blur} illustrates the horizontal motion blur by showing the horizontal ($X$-$Z$) plane of the camera frame, where the vertical motion blur can be interpreted similarly.

The movement of the interest point should be smaller than $1$ pixel for a sufficiently sharp image. 
Interest points captured with a pixel movement greater than $1$ receive a lower score.

\subsubsection{Image Spatial Resolution}\label{subsubsec: res}
The spatial resolution of the image is expressed in millimeter-per-pixel (MMPP), representing the size of the real-world object captured in one image pixel. To achieve a satisfactory resolution for defect inspection, the computed horizontal and vertical resolutions should be smaller than the desired MMPP value. 
Therefore, the resolution metric is computed as:
\begin{align}
    q_\text{res} = \min\left(\dfrac{r_\text{des}}{\max\left(r_\text{horz}, r_\text{vert}\right)}, 1.0\right), \label{eq: resolution metric}
\end{align}
where $r_\text{des}$ is the desirable resolution, $r_\text{horz}$, and $r_\text{vert}$ denote the resolution in the horizontal and vertical image axis, respectively.
We compute spatial resolution in a way similar to the ground sampling distance (GSD) \cite{Wu2024Evaluation} with consideration of the surface normal direction.

\section{Competition at CDC 2023}
The first CARIC competition was held in conjunction with the IEEE Conference on Decision and Control (CDC) in December 2023 in Singapore. The participating teams were required to submit software packages to the organizers for evaluation. The competition featured three distinct scenarios with the following team compositions:

\begin{itemize} \item MBS: Two explorers and three photographers. \item Hangar: One explorer and two photographers. \item Crane: Two explorers and three photographers. \end{itemize}

The number, locations, and sizes of the bounding boxes were unknown to participants before the competition and were provided to the software as parameters at runtime. Interest points, representing areas of potential structural vulnerability, were sampled on the surfaces within the bounding boxes. 
Each scenario had a strict time budget: $600$ seconds for MBS and Crane, and $240$ seconds for Hangar. The time starts to count when any drone takes off, allowing for offline computation before the flight.

The solution of each team was evaluated over five runs, with each test including all three scenarios. Teams were ranked according to the \textit{max} score across the five runs. The score is reported to each team after each evaluation, and the teams are also allowed to update the solution before the deadline. To ensure fairness, all evaluations were conducted on two desktop computers with different hardware configurations: Intel Core i9-13900 CPU with an NVIDIA RTX 4080 GPU, and Intel Xeon W2295 CPU with dual NVIDIA Titan RTX GPUs.

The rankings were consistent across both hardware setups, demonstrating robustness in the evaluation. A total of eight teams submitted software packages, of which seven achieved valid scores while adhering to communication constraints. The top three teams were: \begin{itemize} \item KIOS CoE (University of Cyprus), \item XXH (Nanyang Technological University), and \item STAR (Sun Yat-sen University). \end{itemize}

\section{Winning Team Approaches}
The first CARIC competition showcased diverse approaches to the multi-UAV inspection problem, with teams employing unique strategies for exploration, inspection, and task allocation. The following section presents the approaches of the top three teams, focusing on their methodologies.

\subsection{Team KIOS CoE}
\begin{figure*}
    \centering
    \includegraphics[width=0.8\textwidth]{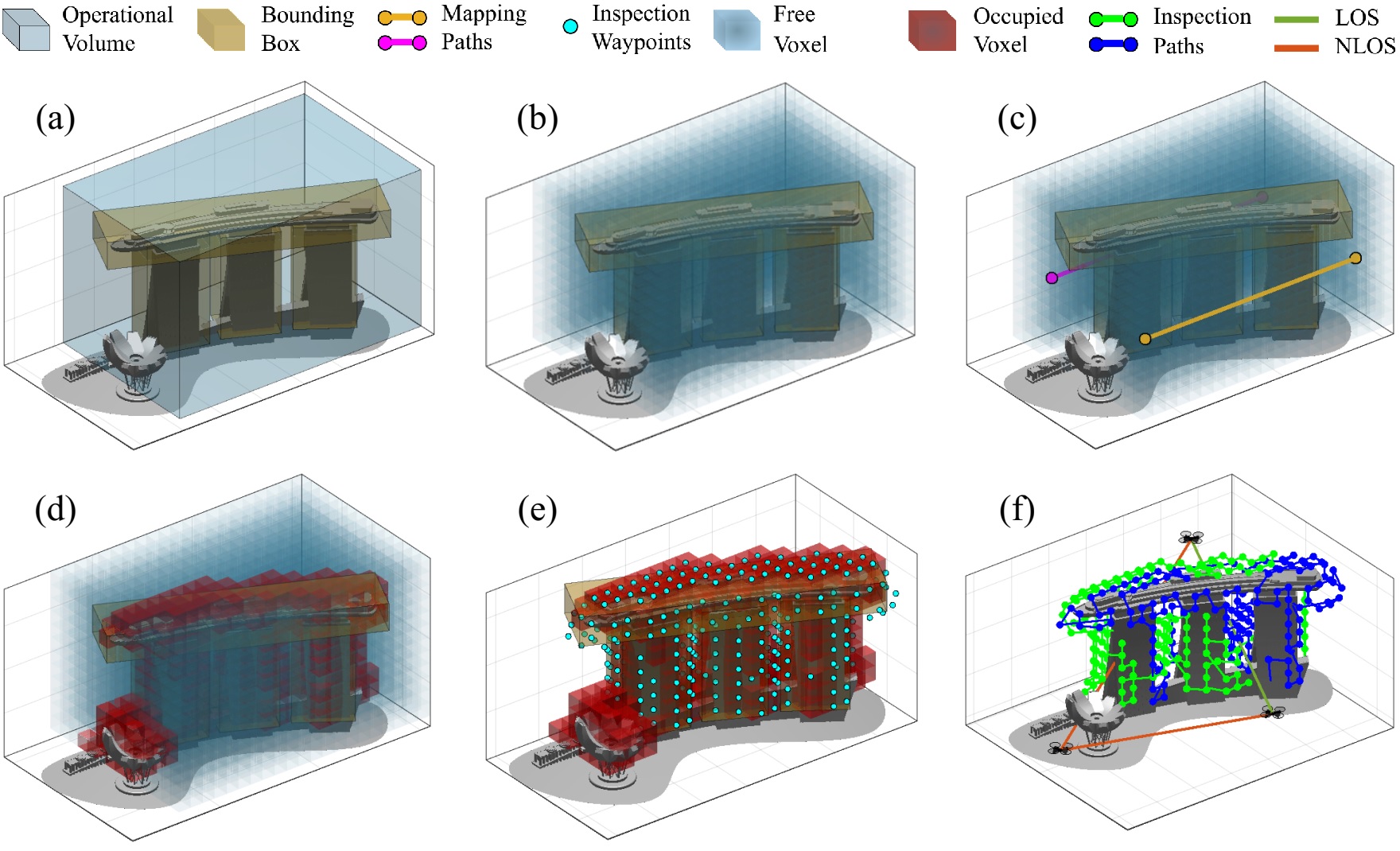} 
    \caption{(a) Derivation of the operational volume, (b) Discretization of the operational volume, (c) Mapping path generation and execution, (d) Initial occupancy map generation, (e) Inspection waypoint generation, (f) Inspection path generation.}
    \label{fig:kios_proposedApproach}
\end{figure*}

\begin{figure}
    \centering
    \includegraphics[clip,trim=0cm 0cm 0cm 0cm,width=\columnwidth]{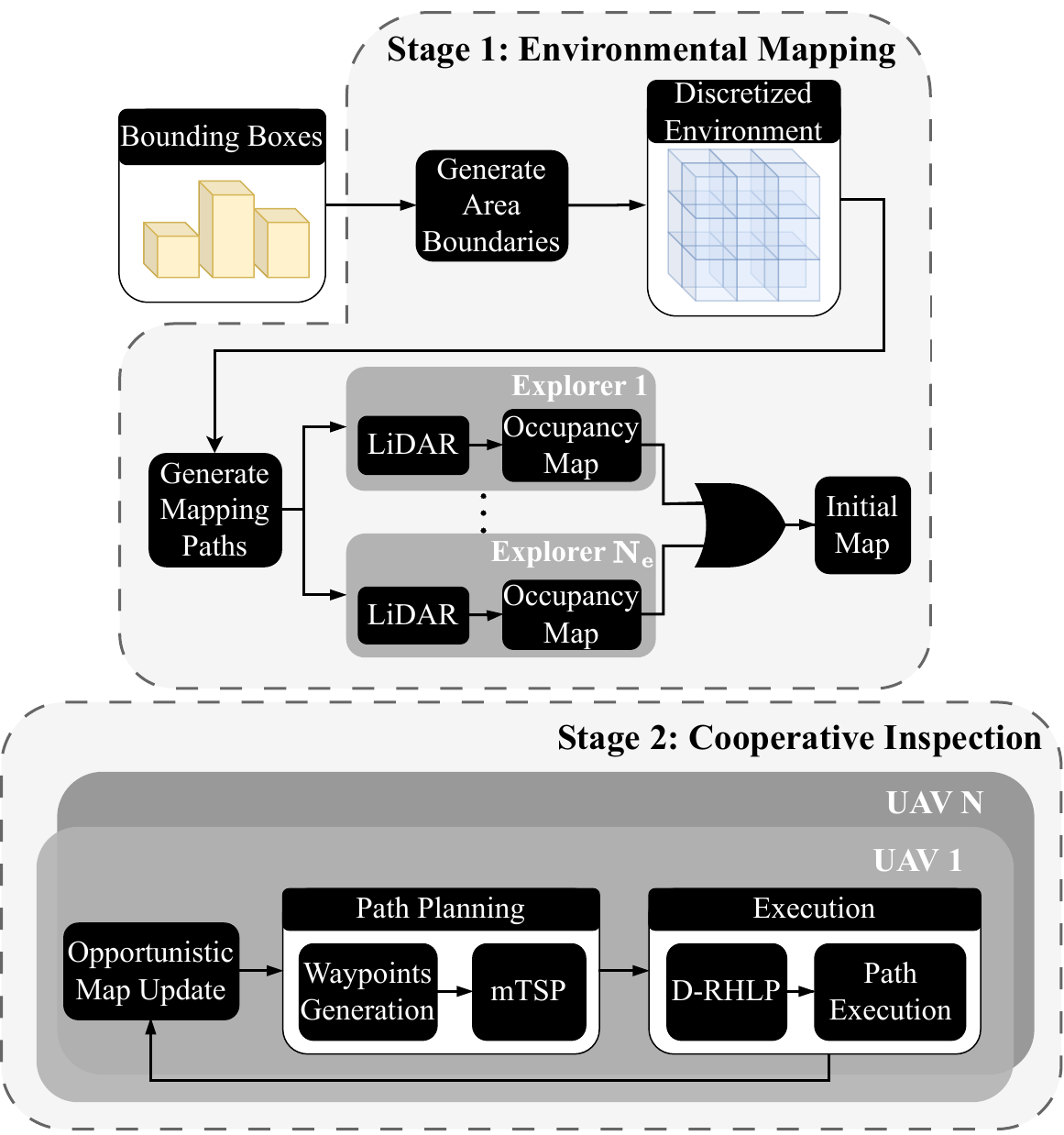}        
    \caption{Overview of the proposed approach for 3D infrastructure inspection using multi-UAV system.}
    \label{fig:kios_overview}
\end{figure}

\subsubsection{Overview}
Team KIOS CoE emerged as the top-performing team in the CARIC competition. 
Their approach is structured into two key stages: \textit{Environmental Mapping} and \textit{Cooperative Inspection}, as illustrated in Fig.~\ref{fig:kios_overview}.
In the first stage, an operational volume $\mathcal{O}$ is defined based on a set of bounding boxes $\mathcal{B}$, which enclose points of interest. The area is discretized into a voxel grid, represented as a graph $\mathcal{G}$, with edges indicating feasible paths (Fig.~\ref{fig:kios_proposedApproach}a-b). Explorers traverse the grid along precomputed paths, using LiDAR to construct occupancy maps. These maps are opportunistically shared with photographers when line-of-sight (LoS) is available, enabling the creation of a global map (Fig.~\ref{fig:kios_proposedApproach}c-d).

In the second stage, UAVs collaboratively inspect operational volume. Inspection waypoints are generated around the occupied voxels (Fig.~\ref{fig:kios_proposedApproach}e). The UAV fleet employs a distributed approach to solve the Multiple Traveling Salesman Problem (mTSP), ensuring efficient path planning. Paths are executed using receding-horizon local planning, dynamically adjusting for real-time occupancy map updates and collision-free operation (Fig.~\ref{fig:kios_proposedApproach}f).

\subsubsection{Environmental Mapping}
The operational volume $\mathcal{O}$ is determined by calculating the smallest cuboid that contains both the infrastructure and the initial positions of the UAVs. This volume is divided into cubic voxels of size $V$, creating a connected graph $\mathcal{G}$, where vertices represent voxels and edges represent possible transitions between them. The adjacency matrix of $\mathcal{G}$ enables efficient navigation and path planning for UAVs.

Explorer UAVs traverse precomputed paths along the longest axis of $\mathcal{O}$, gathering LiDAR data to construct local occupancy maps. Each map identifies voxels that contain obstacles and marks them as occupied. The explorers merge these maps into a global occupancy map and communicate with photographers when there is LoS. This global map provides an initial layout of the operational area, guiding subsequent inspection by excluding paths through occupied voxels.

\subsubsection{Cooperative Inspection}
During the inspection phase, UAVs update their occupancy maps in real time by exchanging data whenever LoS is available. 
Each robot computes the waypoints around the occupied voxels to ensure thorough inspection, where a waypoint is associated with a direction vector to optimize the gimbal angles. 
Each robot computes a subset of the waypoints to travel by solving an mTSP distributively using a method inspired by \cite{anastasiou2020swarm}.
Specifically, each robot generates a set of $N$ paths given the current positions of the robots within LoS and then executes the path tailored to itself.

Paths are executed using a receding-horizon manner, where UAVs dynamically replan routes by solving the mTSP to handle changes in the environment. The adjacency matrix of $\mathcal{G}$ ensures that no two UAVs occupy the same voxel, maintaining collision-free operation. 
While following inspection paths, UAVs adjust their gimbals to focus on specific areas. This iterative process of mapping, waypoint generation, and inspection continues until the mission concludes.


\subsection{Team XXH} 

\subsubsection{Overview}
Team XXH adopts a team-based approach for multi-UAV inspection, grouping UAVs into teams based on their initial positions and roles.
Tasks are allocated among teams proportionally to team size. Each explorer follows a spiral pattern to explore and map assigned regions systematically, while photographers inspect designated layers using similar trajectories.

\subsubsection{Team Formation}
The ground control station (GCS) initializes the process by determining the number of teams. 
Given \(N_e\) explorers and \(N_p\) photographers, with \(N_e < N_p\), the fleet is divided into \(N_e\) teams, each including one explorer responsible for environmental sensing. 
Photographers are assigned to teams based on their proximity to the corresponding explorer.

\subsubsection{Task Assignment} To allocate tasks proportionally among teams, the GCS computes an approximate minimum-length path traversing the longest dimension of all bounding boxes using a best-first search. Each team is then assigned a set of bounding boxes along this path, with the total volume of the assigned regions proportional to the team size. For balanced distribution, bounding boxes may be split along their longest side, dividing regions among teams.

\subsubsection{ Exploration Strategy} Each bounding box is represented as an axis-aligned voxel map, segmented into layers along the longest dimension. The explorer begins exploration at the base layer, following a spiral trajectory around the boundary of the bounding box to identify occupied and unknown voxels. \textbf{A* }search on the voxel map is used to find collision-free paths. Once the region is explored, the explorer transmits the updated map to the photographers. During exploration, photographers remain at the bounding box entry point to maintain line-of-sight communication with the explorer.

\subsubsection{Inspection Strategy} With the updated map, the explorer and photographers collaboratively inspect the occupied regions within the bounding box. The layers are divided among team members, and each UAV scans its designated section using a spiral pattern. The angle of the gimbal is dynamically adjusted to ensure thorough coverage of all exposed voxel faces. To avoid collisions, UAVs share real-time position and trajectory data, with priority rules assigned to prevent deadlocks. Upon completing the inspection of one bounding box, the team transitions to the next assigned region, repeating the exploration and inspection process.

All UAVs exchange position and trajectory data. Obstacle avoidance priorities are assigned based on roles and naming conventions to resolve potential deadlocks. 

\subsection{Team STAR}

\subsubsection{Overview}
\begin{figure}
    \centering
    \includegraphics[width=0.95\columnwidth]{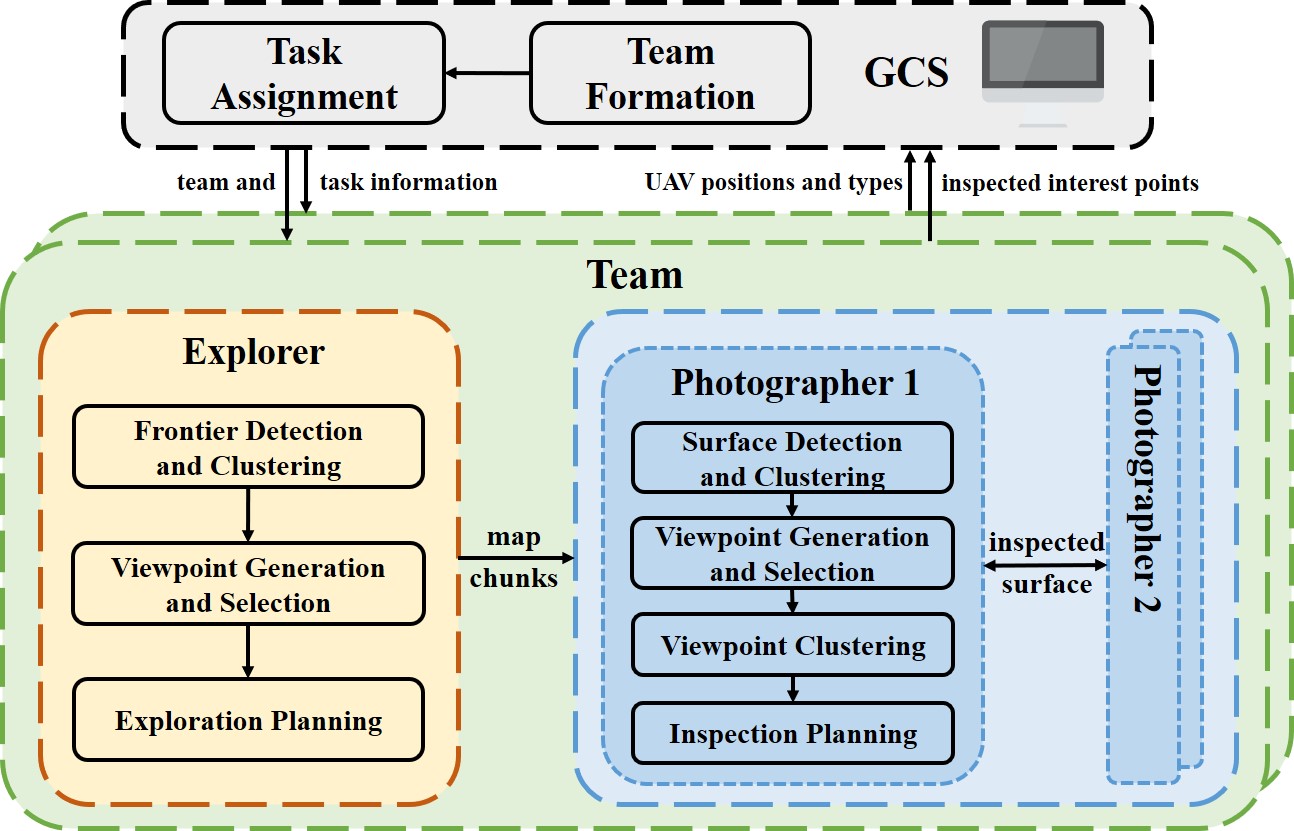}     
    \caption{\label{fig:sys} System overview of Team STAR's method.}
\end{figure}

Team STAR employs a systematic and team-based approach. Each team consists of one explorer and at least one photographer, with tasks allocated by solving an mTSP to optimize workload distribution.
The explorer utilizes the efficient exploration framework FUEL\cite{zhou2021fuel} to explore the environment and transmit the map information to all photographers within the team. 
Photographers then employ a method similar to Star-Searcher\cite{luo2024starsearcher} to inspect the received surfaces. To prevent redundant scans, photographers share their inspected surface information with one another. Concurrently, all inspected interest points are transmitted to the GCS. More details are illustrated in Fig. \ref{fig:sys}.

\subsubsection{Team Formation}
The team formation is similar to the approach of Team XXH, assigning one explorer to each team, and photographers are assigned to teams based on their distance from the explorers. 
The difference is that, to ensure balanced allocation, an upper limit of \(N_t\) photographers per team is set, where \(N_t = \left\lceil {N_{p} / N_{e}} \right\rceil\). 

\subsubsection{Task Assignment}
An mTSP is solved to optimize task assignment among the \(N_e\) teams for \(M\) bounding box task regions. The objective is to minimize the overall travel distance while efficiently assigning inspection tasks to each team. Specifically, a cost matrix for the mTSP solution is constructed such that the cost of inspecting each bounding box is a weighted combination of the distance from the explorer to the box's center and the box's volume.
This cost computation provides an approximation of the workload to facilitate efficient assignment.

\subsubsection{Explorer Strategy}
The exploration framework FUEL\cite{zhou2021fuel} is adopted to acquire environmental information by the explorers. 
Frontiers are detected incrementally and then segmented into appropriately sized clusters using PCA-based clustering. 
Cylindrical sampling regions centered on each frontier are generated to produce viewpoints. 
For each frontier, the viewpoint that maximizes the coverage of frontier cells within the LiDAR field of view (FoV) is selected as the representative viewpoint. An Asymmetric Traveling Salesman Problem (ATSP) is solved based on all representative viewpoints to determine the next best viewpoint to visit. Additionally, when communication with photographers in the same team is available, the explorer transmits map chunks incrementally, similar to the approach used in RACER\cite{zhou2023racer}.

\subsubsection{Photographer Strategy}
Upon receiving the map information from the explorer, the photographer inspects the surfaces on the map and incrementally shares the newly inspected surfaces with other photographers to avoid redundant scanning. Initially, surfaces are detected and clustered using a method similar to the explorer's, and viewpoints are generated and selected accordingly. However, due to the larger FoV of the LiDAR compared to the camera and the higher speed of the explorer relative to the photographer, the photographer often encounters a substantial number of surfaces requiring inspection. Consequently, this results in many viewpoints that need to be planned, leading to significant computational overhead.
To address this issue, the visibility-based viewpoint clustering method and the hierarchical planning approach from Star-Searcher\cite{luo2024starsearcher} are adopted. 
Viewpoints within a distance threshold and without occlusions between them are grouped into clusters. An ATSP is then solved over these viewpoint clusters to provide global guidance. Subsequently, for local planning, another ATSP is solved using the UAV's current position and all viewpoints within the first cluster. 

\subsubsection{Trajectory Planning}
Fast-planner\cite{fast2019zhou} is utilized for all UAVs to generate a smooth and collision-free trajectory from the current position to the next best viewpoint.


\section{Results and Discussion}

\begin{figure*}
    \centering
    \begin{subfigure}[b]{0.24\textwidth}
        \centering
        \includegraphics[trim=0 0 0 0, clip,width=\textwidth]{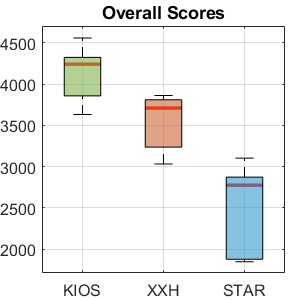}
    \end{subfigure}
    \begin{subfigure}[b]{0.24\textwidth}
        \centering
        \includegraphics[trim=0 0 0 0, clip,width=\textwidth]{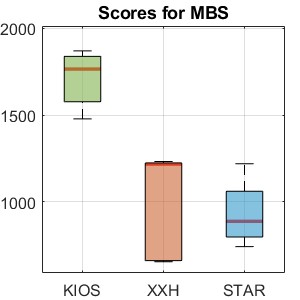}
    \end{subfigure}
    \begin{subfigure}[b]{0.24\textwidth}
        \centering
        \includegraphics[trim=0 0 0 0, clip,width=\textwidth]{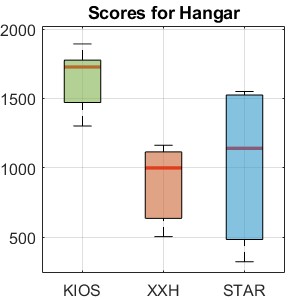}
    \end{subfigure}
    \begin{subfigure}[b]{0.24\textwidth}
        \centering
        \includegraphics[trim=0 0 0 0, clip,width=\textwidth]{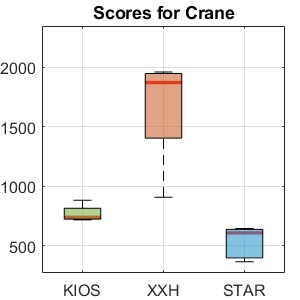}
    \end{subfigure}

    \caption{Box plots of the overall scores obtained by the top three teams across tests, alongside the distribution of scores for individual scenarios: MBS, Crane, and Hangar.}
    \label{fig:scores_boxplot}
\end{figure*}

\begin{figure*}
    \centering

    \begin{subfigure}[b]{0.32\textwidth}
        \centering
        \includegraphics[trim=40 0 60 0, clip,width=\textwidth]{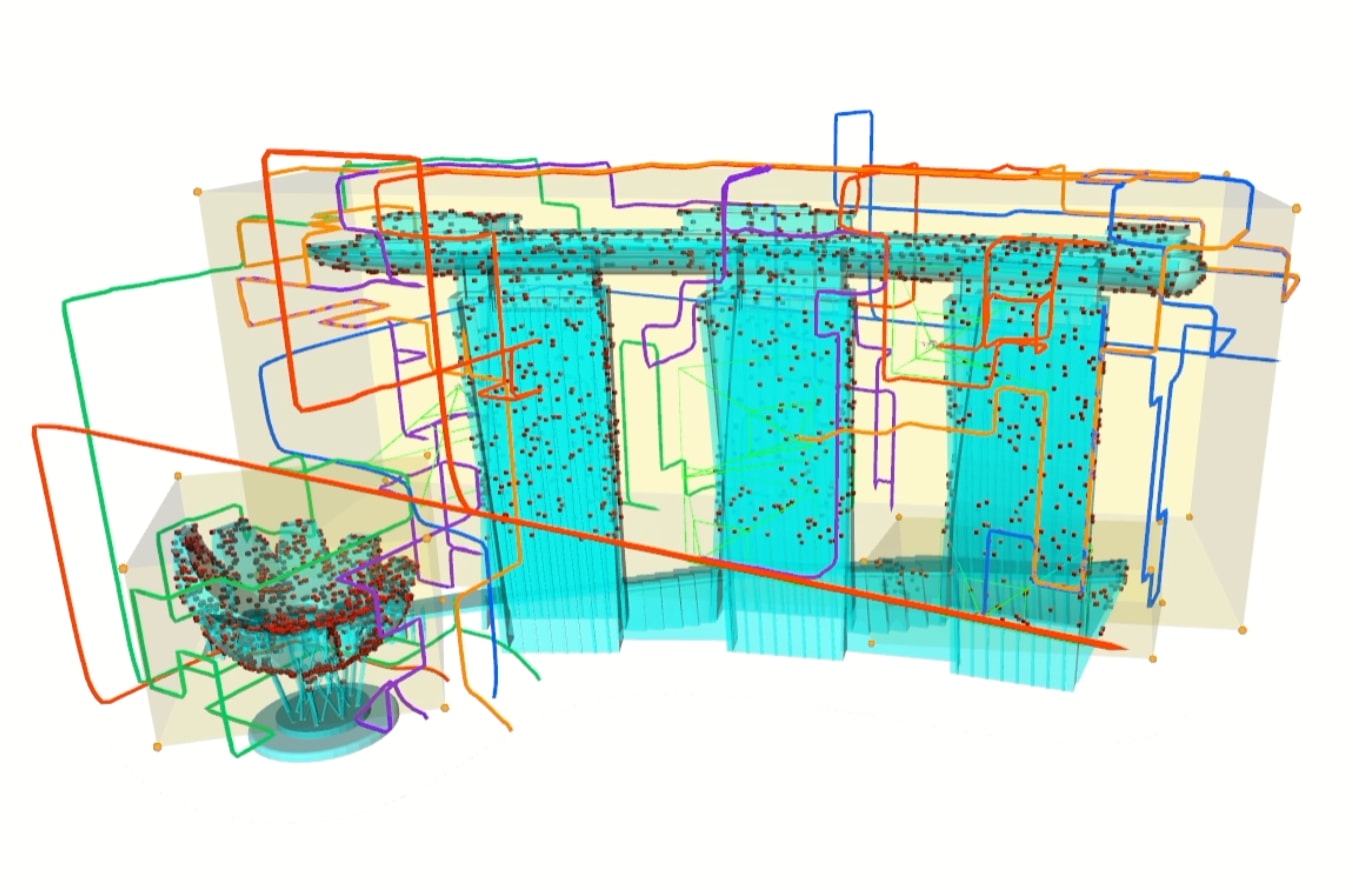}
        \caption{KIOS MBS: $2543$ points detected, $1828$ score}
        \label{subfig: kiosmbs}
    \end{subfigure}
    \begin{subfigure}[b]{0.32\textwidth}
        \centering
        \includegraphics[trim=40 0 60 0, clip,width=\textwidth]{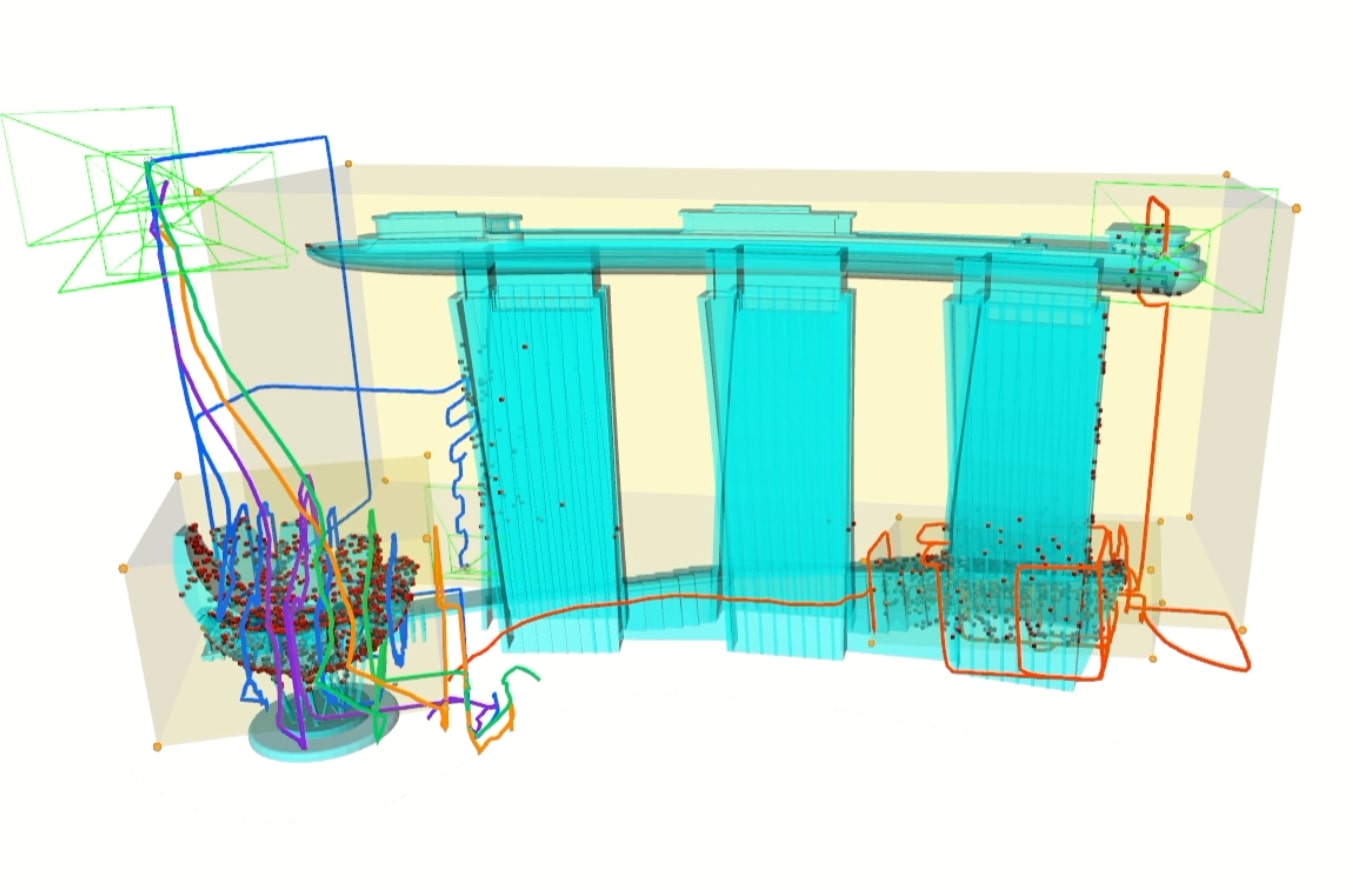}
        \caption{XXH MBS: $1915$ point detected, $1425$ score}
        \label{subfig: xxhmbs}        
    \end{subfigure}
    \begin{subfigure}[b]{0.32\textwidth}
        \centering
        \includegraphics[trim=40 0 60 0, clip,width=\textwidth]{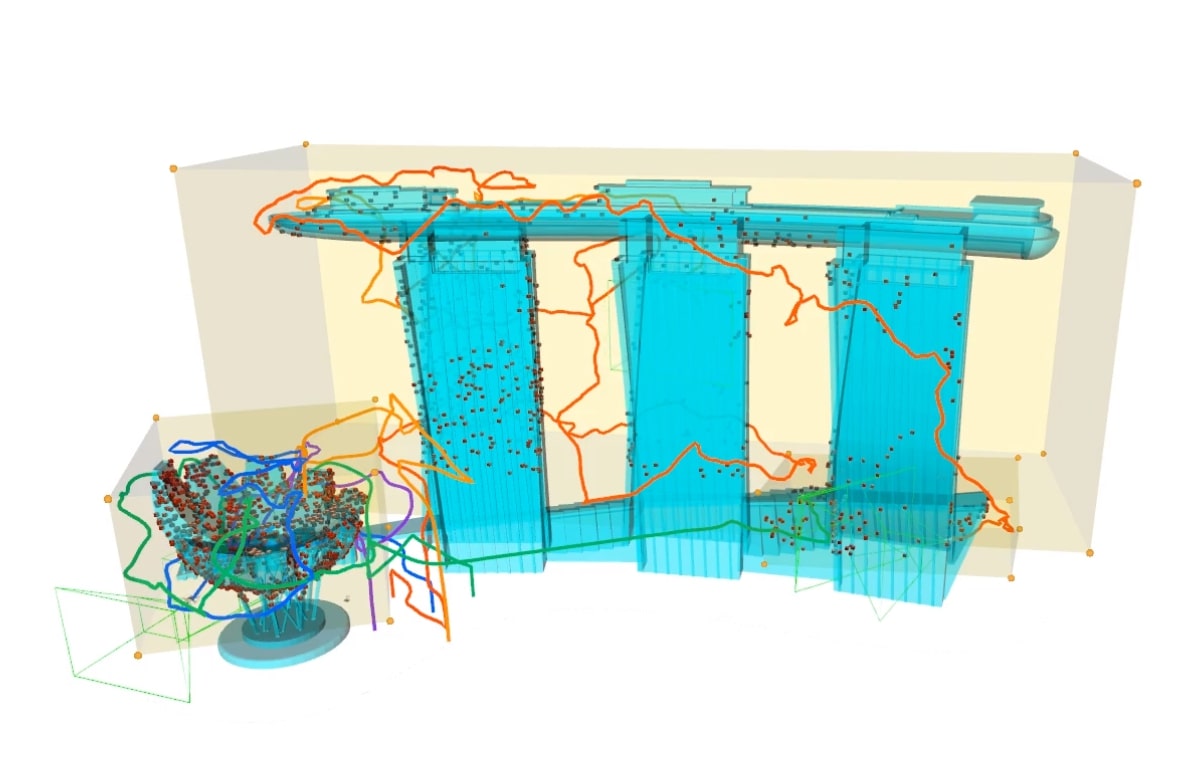}
        \caption{STAR MBS: $1655$ points detected, $1122$ score}
        \label{subfig: starmbs}        
    \end{subfigure}

    \begin{subfigure}[b]{0.32\textwidth}
        \centering
        \includegraphics[width=\textwidth]{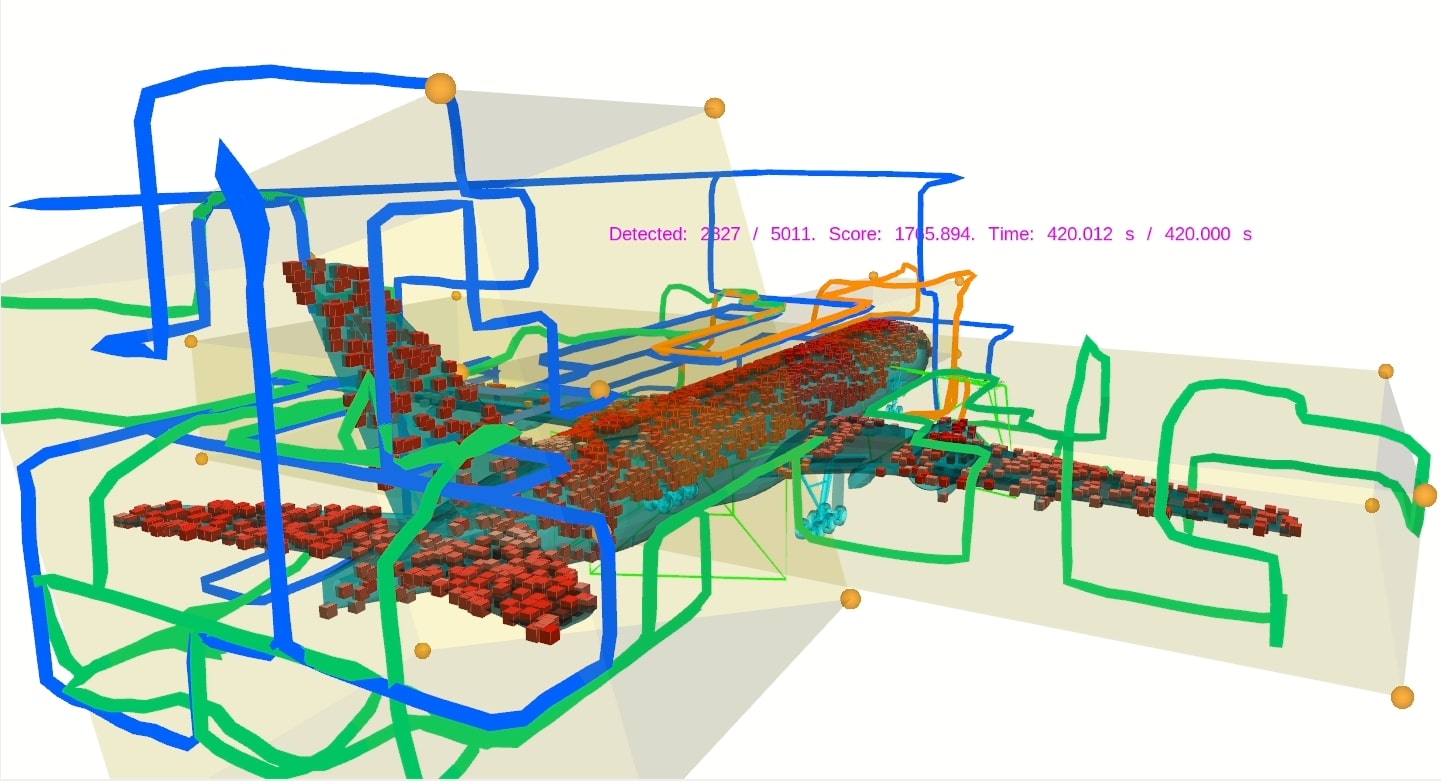}
        \caption{KIOS Hangar: $2827$ points, $1765$ score}
        \label{subfig: kioshangar}        
        
    \end{subfigure}
    \begin{subfigure}[b]{0.32\textwidth}
        \centering
        \includegraphics[width=\textwidth]{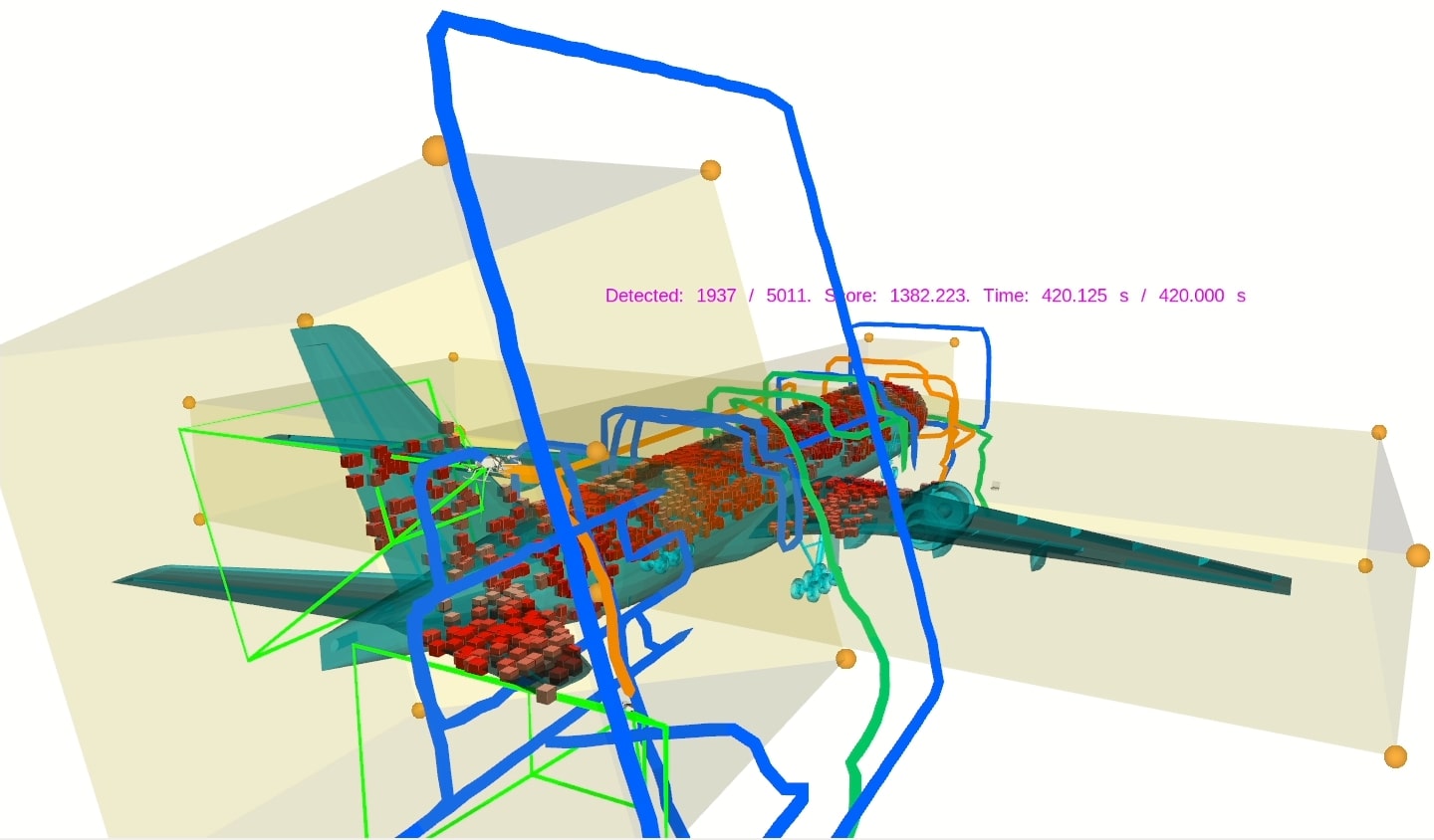}
        \caption{XXH Hangar: $1937$ points detected, $1382$ score}
        \label{subfig: xxhhangar}        

    \end{subfigure}
    \begin{subfigure}[b]{0.32\textwidth}
        \centering
        \includegraphics[width=\textwidth]{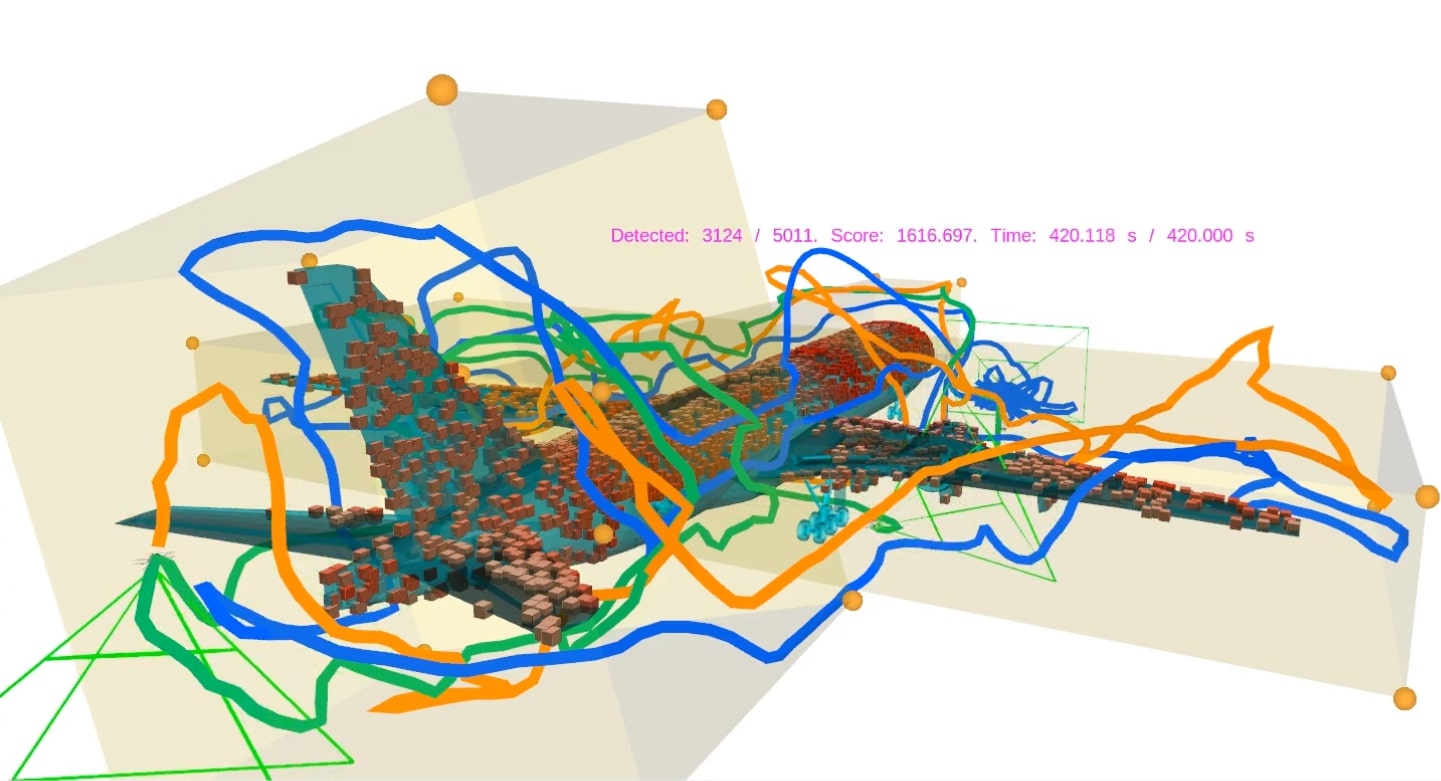}
        \caption{STAR Hangar: $3124$ points, $1616$ score}
        \label{subfig: starhangar}        
        
    \end{subfigure}
    \begin{subfigure}[b]{0.32\textwidth}
        \centering
        \includegraphics[trim=0 0 100 0, clip,width=\textwidth]{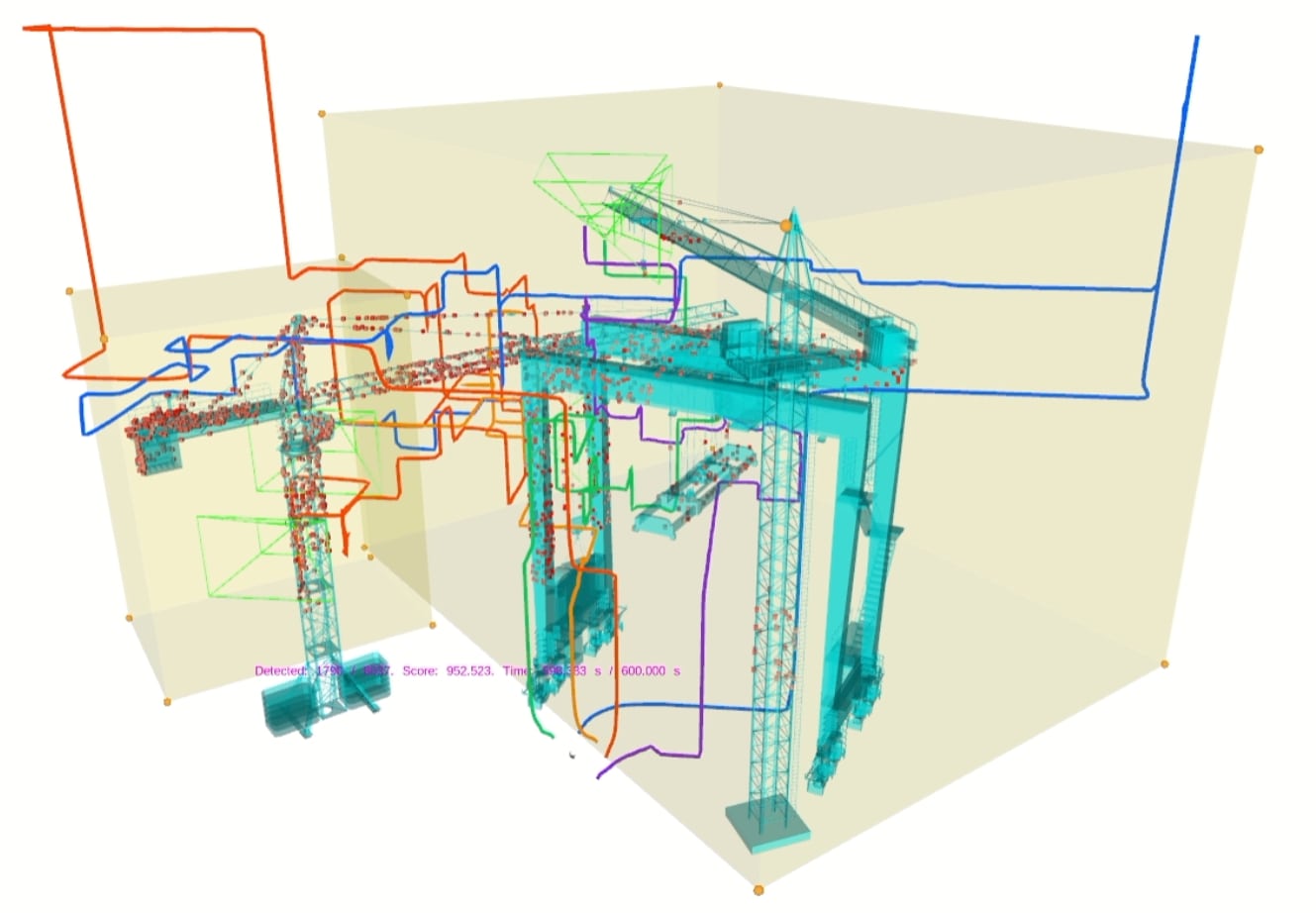}
        \caption{KIOS Crane: $1790$ points detected, $950$ score}
        \label{subfig: kioscrane}        
        
    \end{subfigure}
    \begin{subfigure}[b]{0.32\textwidth}
        \centering
        \includegraphics[trim=0 0 60 0, clip,width=\textwidth]{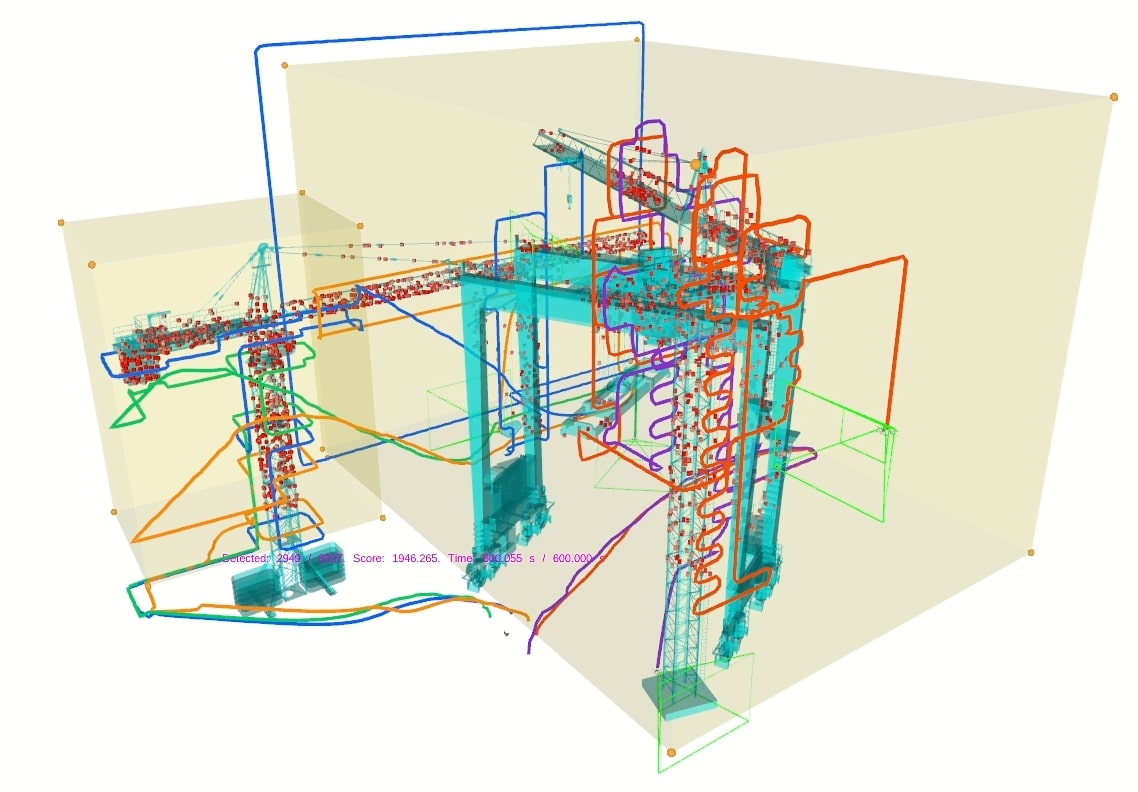}
        \caption{XXH Crane: $2949$ points detected, $1946$ score}
        \label{subfig: xxhcrane}        
        
    \end{subfigure}
    \begin{subfigure}[b]{0.32\textwidth}
        \centering
        \includegraphics[trim=0 0 60 0, clip,width=\textwidth]{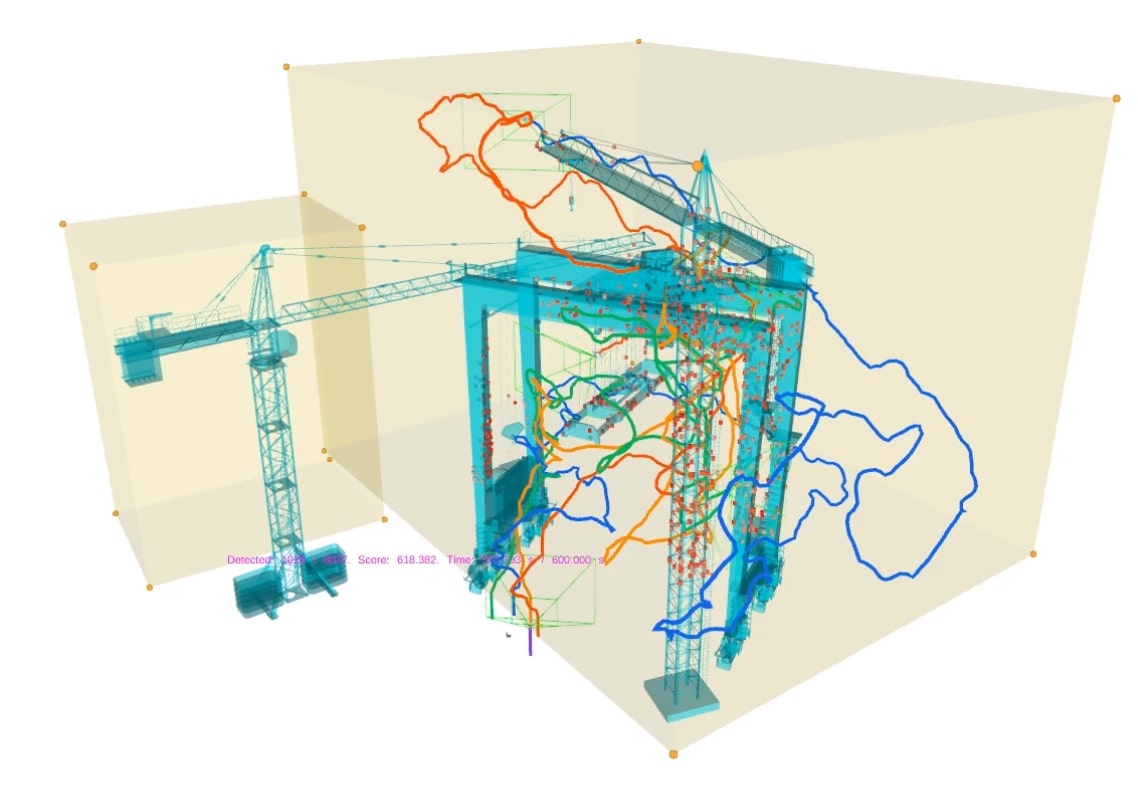}
        \caption{STAR Crane: $1024$ points detected, $619$ score}
        \label{subfig: starcrane}        
        
    \end{subfigure}

    \caption{Illustration of the paths traveled and the points detected (red squares) for each approach. The bounding boxes are shown in transparent yellow.}
    \label{fig:team_comparison}
\end{figure*}

Figure \ref{fig:scores_boxplot} shows the box plots of the overall scores obtained by the top three teams across tests, alongside the distribution of scores for individual scenarios. Figure \ref{fig:team_comparison} illustrates the paths of the UAVs and the final detection results of each scenario in their best-performing runs.
Based on the approaches proposed by the teams and the observation of the teams' performance in the competition, we discuss important lessons from the weaknesses of the methods and highlight potential research directions.

\subsection{Exploration vs Inspection}
KIOS and XXH employ a sequential strategy where exploration precedes inspection, and the photographers stay idle when the explorers initially explore the environment.
However, their approaches differ: 
KIOS performs a rapid exploration along a straight path traversing the longest axis of the operation space containing all bounding boxes; XXH follows a more detailed layer-by-layer spiral path for each bounding box.

Figure \ref{fig:scores_boxplot} shows that KIOS achieves consistent high scores in the MBS and Hangar scenarios, producing the best median and maximum scores. As seen in Figure \ref{subfig: kiosmbs} and \ref{subfig: kioshangar}, KIOS achieves extensive coverage of interest points, including a significant percentage of bounding box regions. 
In contrast, XXH achieves lower scores, leaving parts of the MBS facade (Figure \ref{subfig: xxhmbs}) and the airplane wings and tail (Figure \ref{subfig: xxhhangar}) being omitted from the scan.
As revealed in the video recording of the tests of XXH, the explorers take a long time to conduct the layer-by-layer exploration of a bounding box.
As a result, the photographers only start inspection after a long time, resulting in insufficient time to complete coverage.
In contrast, KIOS's fast exploration strategy allows the team ample time to compute and execute inspection tasks, achieving more coverage.

However, rapid exploration by KIOS can compromise the structural details of the map.
This issue is revealed in the Crane scene, where the thin and closely spaced structures are underrepresented in the map of KIOS, leading to photographers omitting many surfaces during the inspection (Figure \ref{subfig: kioscrane}). 
On the other hand, XXH captures $64\%$ more interest points than KIOS (Figure \ref{subfig: xxhcrane}), demonstrating the importance of detailed exploration in complex environments.

The STAR team adopts a concurrent exploration-inspection strategy, where incremental map updates are immediately shared with photographers. While this approach works well in a constrained environment, such as Hangar (Figure \ref{subfig: starhangar}), it does not achieve good performance in environments with large empty volumes. This is likely because assigning bounding boxes as group tasks is inherently imbalanced, as box sizes can vary significantly, leading to uneven task allocation without a redistribution mechanism. Additionally, frontier-based exploration within large free areas of a box can result in excessive exploration of irrelevant regions, reducing efficiency.

These observations underscore a fundamental trade-off:
rapid exploration maximizes inspection time but risks omitting critical details; 
detailed exploration provides better structural maps but delays inspection.
An incremental exploration-inspection strategy may balance these trade-offs, capturing sufficient structural details without leaving photographers idle. However, such strategies demand sophisticated algorithms to ensure adaptability to diverse environments and compliance with communication constraints.

\subsection{Inspection Quality vs Detection Rate}
KIOS and XXH use waypoint-based navigation, planning one waypoint per grid face to ensure coverage. Photographers briefly stop at each waypoint, adjusting camera angles toward surface normals for precise inspection.
In contrast, STAR optimizes smooth trajectories and continuous camera movement, prioritizing dynamic feasibility and wide coverage.
The continuous yawing motion enables STAR to detect more interest points in scenarios like Crane (Figure \ref{subfig: starcrane}), however, its scores are penalized by low inspection quality.  
This is because the motion blur metric $q_\text{blur}$ deteriorates with a high velocity in the image view (Equation \ref{eq: blur metric}), and the resolution $q_\text{res}$ is affected by the oblique viewing angle (Equation \ref{eq: resolution metric}).
KIOS's stop-and-scan approach achieves a higher overall inspection score despite detecting fewer points.

This trade-off highlights a challenge: increasing the drone speed or camera movement enables faster detection of interest points but with lower quality.
A promising approach is task specialization, where one photographer focuses on rapid detection while another performs slower, high-quality inspection.
Additionally, trajectory planning could incorporate inspection quality metrics to optimize drone speed and camera movement.

\subsection{Task Allocation}
KIOS employs a distributed task allocation strategy where all drones collaboratively solve an mTSP to traverse viewpoints. 
In contrast, XXH and STAR divide robots into teams and assign regions based on bounding box volume and travel distance.
However, this region-based allocation proves problematic, as seen in their performance in MBS (Figure \ref{subfig: xxhmbs} and \ref{subfig: starmbs}). 
In XXH, four drones are assigned to the same team and commanded to inspect the ArtScience Museum (the smaller structure on the left) before moving to the main buildings.
Another explorer is commanded to inspect the smaller bounding box at the bottom right of the main buildings.
Ultimately, the slow pace of the inspection leaves a substantial portion of the facade unchecked.
Similarly, STAR assigns more drones to inspect the ArtScience Museum than the main buildings.

The result shows that a good estimation of inspection workload is critical for a fair allocation of tasks.
A simple assignment strategy based on the volume of the bounding box may cause the robots to spend excessive time on small but complex structures.
In contrast, KIOS's distributed task allocation is based on the viewpoints to inspect the explored map of the structures. This provides a more accurate description of the workload, resulting in more efficient task allocation.

Additionally, separating a small number of robots into multiple teams may not yield efficiency gains compared to a single team due to the difficulty of workload allocation among teams.
However, a single team may become infeasible when the fleet size increases due to the large communication and computation resources.

\subsection{Performance Variance}
\subsubsection{Variance within the Same Scenario} Both XXH and STAR experience significant variances in performance, with best-performing runs exceeding the worst by over 1000 scores in some scenarios (Figure \ref{fig:scores_boxplot}).
In some runs, XXH solution failed to find a feasible path
for explorers, leaving the entire team idle for the whole mission duration.
Investigations reveal that XXH solution has a bug in the waypoints selection logic, ultimately leading to the failure of pathfinding.
Both methods incur collisions during initialization, particularly when drones start from close positions, disrupting subsequent operations.
These issues highlight the need for robust pathfinding algorithms and improved initialization protocols to enhance consistency across runs.

\subsubsection{Variance across Different Scenarios}
No single approach consistently achieves the best performance across all scenarios. KIOS demonstrates superior performance in the MBS and Hangar scenarios but struggles with the Crane scene. Conversely, XXH excels in the Crane scenario, and STAR achieves strong results in Hangar, but both methods underperform in other scenes.
As highlighted in previous sections, these variations stem from differences in exploration strategies, inspection methods, and task allocation approaches, which interact uniquely with the challenges of each scenario. 
These findings underline the importance of developing algorithms that adapt dynamically to different environments and achieve robust performance across various scenarios.





\section{Conclusion}
This paper introduces the Cooperative Aerial Robot Inspection Challenge (CARIC), a benchmark and simulation framework designed to tackle the challenges of heterogeneous multi-UAV inspection in environments without prior structural models. CARIC provides a ready-to-use simulation environment and realistic scenarios to support the development and evaluation of innovative task allocation and motion planning algorithms for heterogeneous UAV teams.

The CARIC CDC 2023 competition showcased diverse strategies for this complex problem. By analyzing the top three teams’ approaches, we highlighted their unique methodologies, trade-offs, and the challenges inherent in the task. The results demonstrated that no single approach has achieved robust performance across all scenarios, emphasizing the difficulty of designing universally effective strategies for cooperative aerial inspection.

Looking ahead, CARIC serves as a foundation for advancing research on cooperative multi-UAV systems. Planned enhancements include expanding the framework with more diverse testing scenarios and transitioning to a fully decentralized implementation using ROS2. We hope the insights and lessons this work presents will inspire continued innovation in this rapidly evolving field.

\bibliographystyle{IEEEtran}
\bibliography{ref, references}

\end{document}